# Chess Player by Co-Evolutionary Algorithm

**Agostinho Rosa**
Laseeb-ISR-IST,
Av Rovisco Pais 1
1049-001 Lisboa, Portugal
acrosa@laseeb.org

**Nuno Ramos**
Laseeb-ISR-IST,
Av Rovisco Pais 1
1049-001 Lisboa, Portugal
nramos@laseeb.org

**Sérgio Salgado**
Laseeb-ISR-IST,
Av Rovisco Pais 1
1049-001 Lisboa, Portugal
ssalgado@laseeb.org

**Abstract** – A co-evolutionary algorithm (CA) based chess player is presented. Implementation details of the algorithms, namely coding, population, variation operators are described. The alpha-beta or mini-max like behaviour of the player is achieved through two competitive or cooperative populations. Special attention is given to the fitness function evaluation. Preliminary test results showed the prove of principle and the program is able to defeat consistently beginner level players and rival experienced one, but it is still not a contender for other computer based implementations

## 1. Introduction

Chess has been since the early steps of electronic computing the most salient mind challenging problem no only for humans but specially for computer scientist and programmers. Artificial Intelligence community and game theory researchers have tried to model the game in order to create strategies and rules for a better understanding of the game quintessence with the final objective of surpassing human players. These developments resulted in the well known defeat of Garry Kasparov (the Word Chess Champion) by IBM's purpose built chess computer "Deep Blue", in 1977 [1]. Very few attempts have been made to address two players' games by evolutionary algorithms. For the Go-moku game a genetic algorithm (GA) based program is described in [2]. For the checkers game an evolutionary based search program [3]. For the Chess game there is only a few experiments have been described [4]. Co-evolutionary strategies have been applied to Backgammon [5] and iterated prisoner dilemma [6]. Since chess is a well-known game and there are many references describing the rules of the game and also different type of machine intelligence solutions [7]. In this paper we restrict to the presentation of the implementation aspects of the co-evolutionary algorithm based chess machine player.

## 2. Game

Chess game is a 2 player strategic game played in an 8x8 "chess" board (alternating black and white squares). Each player has the same set of pieces (8 Pawns, 2 Knights, 2 Bishops, 2 Rooks, 1 Queen, and 1 King); the different pieces have different movement patterns. The objective is to take the opponent king (check mate). Each player makes their moves alternatively [7].
The search space in a game of chess problem is NxM, where N is number of possible choices and M the depth level (number of look ahead moves) is in average (N=35 and M=4) will ends up to 1500625 choices.

## 3. Methodology

The chess player could be implemented as a usual in evolutionary algorithms (EA), where the population represents candidate sequences of alternated (white and black) moves. Another possibility is to use two different populations, where each element of the population is a list of moves of only one of the opponents, black or white. The situation is usually known as co-evolution, where more than one population evolves together with specific form of interaction.
The first option is simpler to implement but needs a very large population in order to cover a representative number of play sequences. The advantage of the second is a more compact representation of the moves and also provides a finer control on the number of play sequences to analyse. If K is the size of each of the two equal size populations, then we can obtain KxK possible combinations of alternated play sequences. Different strategies can be used to reduce the number of evaluations, like for example the most promising ones.

### 3.1 The co-evolutionary algorithm

The CA is an EA with two distinct populations, one for black and one for white. The EA used for each population is the standard binary coded GA with fitness proportional selection with elitism, crossover and variation operators. The variation operators, crossover and mutation, are applied to these two populations independently, obtaining two offspring populations. The fitness of each individual is calculated in each generation, takes into account not only the quality of individual moves in his own population but also the quality of the possible moves of the other population. For example, a move that takes a knight but loses the queen in the next move is not a god trade-off.

### 3.2 Population and coding

The two populations have the same number K of individuals; there is no specific reason to make them different. The size of the population depends on the depth level of the moves analysed in order to maintain a suitable percentage of coverage of the search space.
Each individual is coded by a binary chromosome of variable number of genes, as shown in figure 1. The number of genes is the depth level or the number of the play-ahead moves.
The genes are binary codes, where length and coding depends on the specific piece. Each gene represents a possible move and contains the information of the piece type, the move and the distance of the move. The generic gene structure is shown in figure 2.

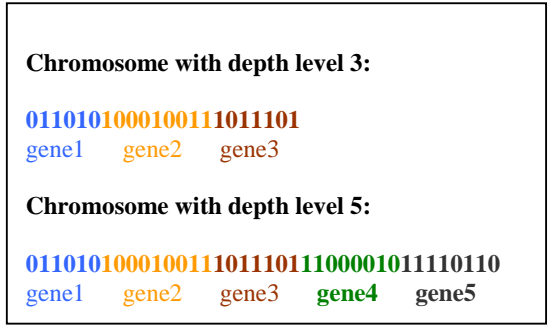

Figure 1 – Chromosomes of varying length dependent on the depthlevel.

| Piece (4 bits) | Direction (3 bits) | Displacement (3 bits) |
|---|---|---|

Figure 2 – Gene coding: Type of Piece, Move Direction and Displacement.

The Piece code length is 4 bits representing all the 16 pieces in the game. Table 1 shows the coding used.

| Code | Piece |
|---|---|
| 0000 | Pawn 1 |
| 0001 | Pawn 2 |
| 0011 | Pawn 3 |
| 0010 | Pawn 4 |
| 0110 | Pawn 5 |
| 0111 | Pawn 6 |
| 0101 | Pawn 7 |
| 0100 | Pawn 8 |
| 1000 | Rook 1 |
| 1001 | Rook 2 |
| 1010 | Knight 1 |
| 1011 | Knight 2 |
| 1100 | Bishop 1 |
| 1101 | Bishop 2 |
| 1110 | Queen |
| 1111 | King |

Table 1 – Type of piece Coding

The Direction coding depends on the type of piece. The Pawns, Rooks, and Bishops have only 4 different directions; Knights and Queen have 8 and the King has 10 (8 for direction and 2 for Left and Right castle). Therefore, a maximum of 4 bits is needed for the Direction coding.
The displacement is the number of squares a piece can move in any direction. Pawns, Knights and King have a fixed displacement of 1, only Rooks, Bishops and Queen needs the displacement code.
In order to have a more compact code and avoiding redundancy, a variable size coding is used. For the Pawns a total of 6 bits is needed and a total of 10 bits for the Queen.

### 3.3 Variation Operators
The Variation operators used are: bit-level uniform crossover operator with probability 0.7 and bit mutation and/or simple inversion with probability 0.02 per bit.

### 3.4 Repair function
The Variation operators can make the chromosomes invalid. Since each piece has its own specific coding, crossover and mutation can change any or all the gene code fields. In order to avoid wasting computation time, invalid chromosomes go through a repair function after the application of the variation operators, before fitness evaluation. The repair function not only corrects invalid chromosomes but also detects pieces that are no longer in play.

### 3.5 Evaluation Strategy
As mentioned before, the fitness function evaluation is the heart of the player intelligence.
A new set of chromosomes are formed through the combination of a pair of black and white chromosomes. Each chromosome is formed by alternated white and black genes, as shown in figure 3.

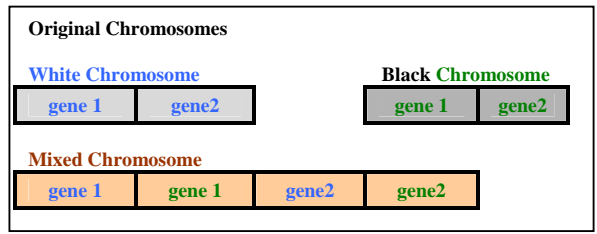

Figure 3 – Mixed chromosomes, a combination of a pair of white and black chromosomes.

The new population is evaluated by a static fitness function and the fitness of best elements of the two populations is elaborated further through the mixed chromosomes. The P best white chromosomes are combined with the best Q black chromosomes, resulting in PxQ mixed chromosomes.

The first move of white chromosome with the best mixed chromosome fitness is played. The corresponding first gene of both, black and white chromosomes is discarded and a new randomly generated gene is appended at the end.

### 3.6 Fitness function
Each piece in the game has a relative weight factor, absolute and relative positional (AP and RP) and menace-protection (MP) scorings. The relative weight is dependent on the relative value given to the different pieces in the game. There are several proposals for the relative weight and some are even optimized by GA through simulated game plays [4]. Here an empirical weight system was adopted, and it is similar to most often adopted ones, as shown in table 2.

| Piece | Weight |
|---|---|
| **Pawn** | 100 |
| **Knight** | 300 |
| **Bishop** | 320 |
| **Rook** | 500 |
| **Queen** | 900 |
| **King** | 3000 |

Table 2 – Relative weight factor for the pieces.

The absolute positional scoring is the corresponding value an 8x8 weight matrix, it depends only on the position of the piece in the game board. It reflects the strategic positional value of the piece and is dynamic along the game.

The relative positional scoring takes into account of the synergetic value of the interaction of pieces when they are close together.

The menace-protection scoring depends on the balance value between the number of pieces protecting a specific piece and the number of menacing pieces from the opponent. When a piece is under menace the MP scoring is calculated by, subtracting the value of the menaced piece, adding the value of the attacked piece and subtracting the value of the protected piece. An example is provided in figure 4; the Black Knight is under the menace/attack of 3 white pieces and is protected by only 2 black pieces. If the last move was the Black Knight then the MP scoring of the Black knight only will be: subtract Black Knight (-300), add White Knight (300), subtract Black Bishop (-320), add White Bishop (320) and subtract Black Queen (-900). The total MP scoring will be -900.

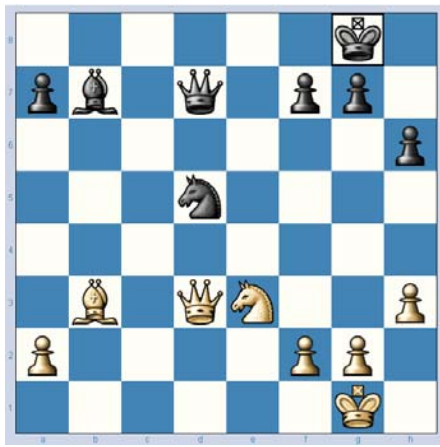

Figure 4 – Menace-protection of a piece

*3.6.1 Pawns*
The pawn has several absolute positional scoring tables. In the beginning of the game, the pawns have very limited value; on the other at the final phases, the pawns have a determinant role in outcome of the game. The pawns AP scoring table also changes after Right or Left castle. Table 3 and 4, shows the Pawn AP scoring matrix in the beginning and end of the game respectively.

| 0 | 0 | 0 | 0 | 0 | 0 | 0 | 0 |
|---|---|---|---|---|---|---|---|
| 0 | 0 | 0 | 0 | 0 | 0 | 0 | 0 |
| 5 | 10 | 15 | 20 | 20 | 15 | 10 | 5 |
| 4 | 8 | 12 | 16 | 16 | 12 | 8 | 4 |
| 3 | 6 | 9 | 12 | 12 | 9 | 6 | 3 |
| 2 | 4 | 6 | 8 | 8 | 6 | 4 | 2 |
| 1 | 2 | 3 | 4 | 4 | 3 | 2 | 1 |
| 0 | 0 | 0 | 0 | 0 | 0 | 0 | 0 |

Table 3 – Pawn beginning AP Scoring Matrix

In the beginning stage of the game the centre positions on the board have larger strategic score in terms of protection and menace. At the later stage of the game positions closer to the final line has greater value, since it can become any piece of choice.

| 20 | 30 | 40 | 50 | 50 | 40 | 30 | 20 |
|---|---|---|---|---|---|---|---|
| 12 | 24 | 36 | 48 | 48 | 36 | 24 | 12 |
| 10 | 20 | 30 | 40 | 40 | 30 | 20 | 10 |
| 8 | 16 | 24 | 32 | 32 | 24 | 16 | 8 |
| 6 | 12 | 18 | 24 | 24 | 18 | 12 | 6 |
| 4 | 8 | 12 | 16 | 16 | 12 | 8 | 4 |
| 2 | 4 | 6 | 8 | 8 | 6 | 4 | 2 |
| 0 | 0 | 0 | 0 | 0 | 0 | 0 | 0 |

Table 4 – Pawn Final stage AP Scoring Matrix

The Left and Right Castle also changes the AP scoring Matrix, according to specific position of the pieces after the Castle.

The Relative Positional scoring of the Pawns is the following:
- Add 3 points for each protecting Pawn
- Subtract 7 points for each Doubled Pawn (Pawns in the same column).
- Subtract 3 points, if the Pawn is isolated.
- Subtract 10 points for each Pawn in columns without opponent Pawns.
- Add 15 points the Passed Pawns, (there is no opponent Pawns in the same, immediate left and right columns).
- Add 10 points for linked and passed Pawns, (besides been passed the Pawn is also protected by another Pawn of the same colour).
- Subtract 7 (or 3) points for each passed and blocked Pawn by a Knight (or Bishop), (a passed Pawn cannot move, because there is an opponent Knight or Bishop occupying the square in front of it.

The Doubled Pawn is shown in figure 5 at position C7. An Isolated white Pawn at D4 is shown in figure 6.

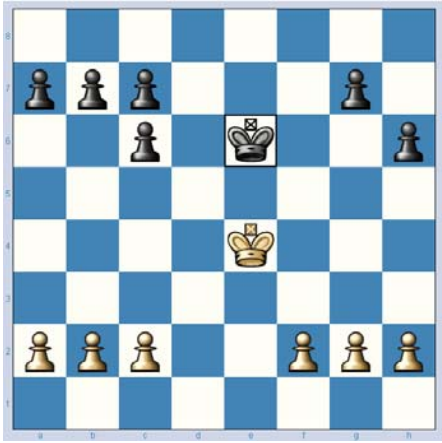

Figure 5 – Doubled Back Pawn at C7

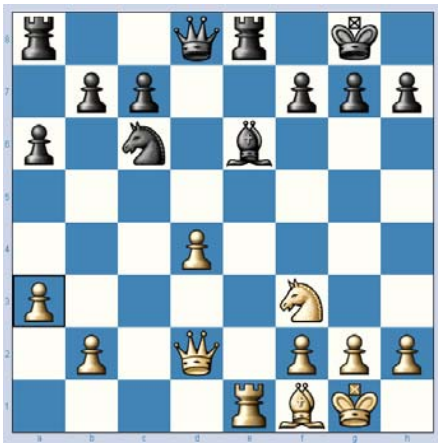

Figure 6 – An Isolated White Pawn at D4

A Passed and Blocked white Pawn at E5 by a Black Knight at E6 is shown in figure 7.

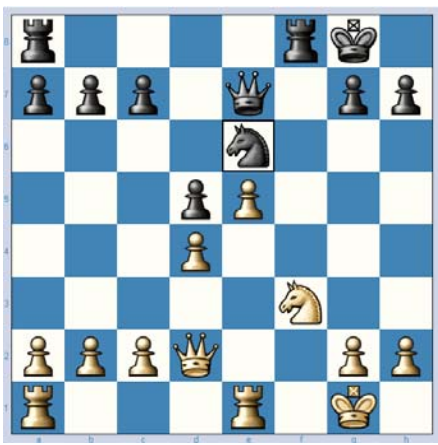

Figure 7 – Passed and Blocked white Pawn at D5 by a Black Knight at D6.

### 3.6.2 Rooks

The two Rooks have a distance action and can protect each other, so the absolute position is not important; the AP scoring is substituted by The Proximity and Mobility Scorings. The Mobility scoring is shown in table 5. Mobility is the total number of squares each Rook can move to.

The Proximity Score of the Rook is shown in table 6. Proximity is the sum of column and row distances of the Rook to the opponent King's position. The reason of this scoring is due to the movement restriction inflicted to the opponent King. For example a Rook at the distance of 1 Row and 2 Columns will add 24 points (14 +10).

The RP scoring is the following:
- Add 20 points for each Rook of the same colour present in line 7 (or 2), as shown in figure 8.
- Add 15 points for the presence of 2 or more Rooks in the same column.
- Add 3 points if opponent Pawns are under menace.
- Add 4 points for absence of opponent Pawns in the same column.
- Subtract 12 points, if the King's Rook is moved before the King. (Will disable the Left Castle)
- Subtract 8 points, if the Queen's Rook is moved before the King. (Will disable the Right Castle).

| Rook Mobility | Scoring |
|---|---|
| 0 | -4 |
| 1 | -3 |
| 2 | -2 |
| 3 | -1 |
| 4 | 0 |
| 5 | 1 |
| 6 | 2 |
| 7 | 3 |
| 8 | 4 |
| 9 | 5 |
| 10 | 6 |
| 11 | 6 |
| 12 | 6 |

Table 5 – Rook Mobility Score

| Rook Proximity | Scoring |
|---|---|
| 1 | 14 |
| 2 | 10 |
| 3 | 8 |
| 4 | 5 |
| 5 | 3 |
| 6 | 1 |
| 7 | 0 |

Table 6 – Rook Proximity Scoring

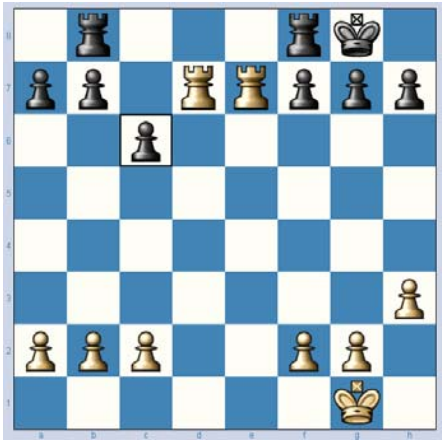

Figure 8 – Two white Rooks, present at row 7 (add 40 points).

*3.6.3 Knights*
The AP scoring of the Knight is shown in table 7. The movements of the Knights are more restricted at the edges of the board than at the centre. Figure 9 shows an example of a Free and blocked Knight.

| -10 | -5 | -5 | -5 | -5 | -5 | -5 | -10 |
|---|---|---|---|---|---|---|---|
| -5 | 0 | 0 | 0 | 0 | 0 | 0 | -5 |
| -5 | 0 | 5 | 5 | 5 | 5 | 0 | -5 |
| -5 | 0 | 5 | 10 | 10 | 5 | 0 | -5 |
| -5 | 0 | 5 | 10 | 10 | 5 | 0 | -5 |
| -5 | 0 | 5 | 5 | 5 | 5 | 0 | -5 |
| -5 | 0 | 0 | 0 | 0 | 0 | 0 | -5 |
| -10 | -5 | -5 | -5 | -5 | -5 | -5 | -10 |

Table 7 – Knight AP Scoring Matrix

Table 8 shows the Mobility Scoring of the Knights. Mobility is the number of empty squares that the Knight can move to.

| Knight Mobility | Scoring |
|---|---|
| 0 | -6 |
| 1 | -2 |
| 2 | 1 |
| 3 | 2 |
| 4 | 3 |
| 5 | 4 |
| 6 | 5 |
| 7 | 6 |
| 8 | 7 |

Table 8 – Knight Mobility Scoring

The Proximity Scoring is shown in table 9.

| Knight Proximity | Scoring |
|---|---|
| 1 | 12 |
| 2 | 10 |
| 3 | 8 |
| 4 | 6 |
| 5 | 4 |
| 6 | 2 |
| 7 | 0 |
| 8 | 0 |
| 9 | -1 |
| 10 | -2 |
| 11 | -3 |
| 12 | -4 |
| 13 | -5 |
| 14 | -6 |

Table 9 – Knight Proximity Scoring

The RP Scoring of Knight is:
- Add 3 points for each protecting Pawn to Knights at a proximity lower than 7.

*3.6.4 Bishops*
The AP scoring of the Bishop is shown in table 10. It is very similar to the Knight AP Scoring. As reflected in the score differences, the movement restriction of the Bishop is less severe at the board edges than for the Knight. Figure 10 shows an example of free (D3) and blocked (D7) Bishops.

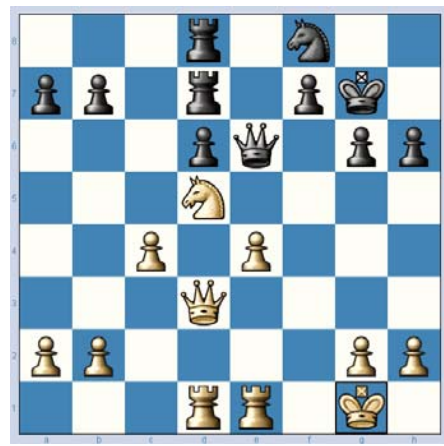

Figure 9 – Freedom of White Knight at D5 and Blocked Black Knight at F8.

| -1 | -5 | -3 | -5 | -5 | -3 | -5 | -1 |
|---|---|---|---|---|---|---|---|
| -3 | 10 | 0 | 10 | 10 | 0 | 10 | -3 |
| -1 | 3 | 6 | 10 | 10 | 6 | 3 | -1 |
| -1 | 10 | 10 | 3 | 3 | 10 | 10 | -1 |
| -1 | 10 | 10 | 3 | 3 | 10 | 10 | -1 |
| -1 | 3 | 6 | 10 | 10 | 6 | 3 | -1 |
| -3 | 10 | 0 | 10 | 10 | 0 | 10 | -3 |
| -1 | -5 | -3 | -5 | -5 | -3 | -5 | -1 |

Table 10 – Bishop AP Scoring Matrix

| Bishop Mobility | Scoring |
|---|---|
| 0 | -4 |
| 1 | -3 |
| 2 | -2 |
| 3 | -1 |

| 4 | 0 |
| 5 | 1 |
| 7 | 3 |
| 8 | 4 |
| 9 | 5 |
| 10 | 6 |
| 11 | 6 |
| 12 | 6 |
| 13 | 6 |

Table 10 – Bishop Mobility Scoring

The RP scoring of the Bishop is:
- Add 20 points, if both Bishops of the same colour are present. (Bishops are complementary, each acting on black or white squares exclusively).
- Subtract 3 points for each Pawn (independent of colour) present in the adjacent diagonal. (The Bishops loose its effectiveness when obstructed).

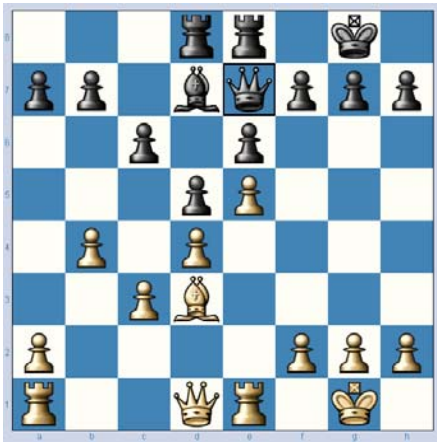

Figure 10 – Free white Bishop (D4) and blocked Black Bishop (D7).

### 3.6.5 Queen

The Queen as the Rooks do not have AP scoring matrix.
The Mobility can be very large, the upper limit is 28. Two different matrixes are used for the beginning and end stages of the game, reflecting the increased importance of the queen, when there are few pieces in play.
The Proximity scoring of the Queen is shown in table 11.

| Queen Proximity | Scoring |
|---|---|
| 1 | 35 |
| 2 | 27 |
| 3 | 21 |
| 4 | 15 |
| 5 | 11 |
| 6 | 8 |
| 7 | 6 |
| 8 | 5 |
| 9 | 4 |
| 10 | 3 |
| 11 | 2 |
| 12 | 1 |
| 13 | 0 |
| 14 | 0 |

Table 11 – Proximity Scoring of the Queen.

The Proximity scoring is very high, especially for small values of proximity. When the Queen is very close to the opponent King it restricts drastically its movements,
The RP scoring of the Queen is:
- Add 9 points for the presence of a Bishop in the same diagonal occupied by the Queen. (A protected Queen is a serious menace for the opponent King).
- Subtract 9 points, if the Queen is moved before two minor pieces (Knight or Bishops). (The Queen is a very powerful and valuable piece, should not be too exposed prematurely).
- Add 6 points if the Queen is on the row 7 (or 2)
- Add 6 points if the column of the queen is free from any Pawn.

### 3.6.6 King

The King is the most valuable piece in the game, there is no widely accepted weighting and scoring values, but it is of general consensus that it should at least be more than the some of all other pieces.
The RP Scoring of the King is:
- Subtract 10000 points if suffer check-mate
- Add 30 points if Castle
- Subtract 30 points if the first move of the King is not a Castle.
- Add 10 points for each piece difference of friendly and foe pieces surrounding the King (the Queen counts here as 3 pieces).
- Subtract 10 points for each movement of protecting pawns after Castle.

There is also two AP scoring for the beginning and end stages of the game for the King. During the beginning of the game a well protected and covered positions are rewarded but advanced positions are highly penalized, as shown In table 12. At the end stages the centre of the board has more strategic value.

| -35 | -35 | -35 | -35 | -35 | -35 | -35 | -35 |
|---|---|---|---|---|---|---|---|
| -30 | -30 | -30 | -30 | -30 | -30 | -30 | -30 |
| -25 | -25 | -25 | -25 | -25 | -25 | -25 | -25 |
| -20 | -20 | -20 | -20 | -20 | -20 | -20 | -20 |
| -15 | -15 | -15 | -15 | -15 | -15 | -15 | -15 |
| -10 | -10 | -10 | -10 | -10 | -10 | -10 | -10 |
| 0 | 0 | -3 | -5 | -5 | -3 | 0 | 0 |
| 5 | 10 | 10 | 0 | 0 | 5 | 10 | 5 |

Table 12 – The King AP Scoring at the beginning of the game.

Figure 11 shows an example of protected white King and unprotected black King.

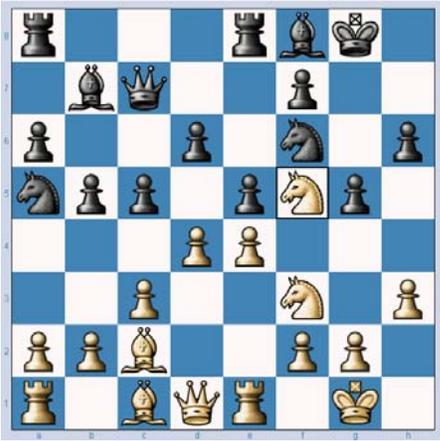

Figure 11 – Protected white King and unprotected black King.

*3.6.7 End stages*
The end stage threshold condition is the presence of less than 6 minor pieces (Knights and Bishops) in the game.

*3.6.8 Technical Tie and Checks*
A Technical Tie condition is declared when one of the following conditions is met:
 - King against King
 - King and Knight against King
 - King and Bishop against King
 - King and Bishop against King and Bishop
 - King and Bishop against King and Knight
 - King and 2 Knights against King
 - Repetition of the same last 3 moves by both players.

Technical Tie is not possible at presence of any Queen, Rooks or Pawn still in play. The same also applies when more than 2 Knights or Bishops are present in the game. The technical tie check is performed whenever a piece is taken.
For a more detailed description of the calculation and scoring procedure of fitness function, see [8].

Before each move is executed, all the rules are checked first. Forbidden moves like exposing the King to check or signalling a check situation to the opponent will be performed. Another situation detected is the full blocking (there is no valid move) where the defeat is awarded to the blocked player. The Stalemate is also detected (when the only valid moves will expose the King to check, situation in which a defeat is awarded.

*3.6.9 Implementation aspects*
The CA Chess player satisfies all the internal rules of Chess, namely the Pawn *empassant move* (when a Pawn steps 2 squares in the first move and cross adjacent columns opponent pawns, the Pawn can be taken by the opponent Pawn as if only one square has been moved) and the Pawn promotion (when a Pawn reaches the last row in the opponent side it is promoted to any piece of choice, except the King and Pawn. The CA Chess Player automatically chooses the Queen, which is the piece of choice, except very rare situation, where a different piece could be chosen. King Left (or Right) Castle is a complex move where the King (Queen) Rook and the King exchange positions simultaneously.

The program is implemented using Java 2; it is available by request through the authors. In a Pentium IV 1 GHz the average time for a move using the default configuration is 10 seconds.

## 4 Results

Two sets of tests are done and presented here. The first is algorithm vs. algorithm. These tests aim to observe the behaviour of different CA settings. In the different configurations a fair comparison in terms of computation time is tried, but due to the characteristics of the algorithm it is difficult to ensure for tests 4.1.1. Besides the differences in the algorithms, its stochastic nature and the uncertainty of the repair function will make the computation different on every run. The second test is algorithm vs. human players; it aims to classify the performance of the CA against different level of human players.

*4.1 Algorithm vs. Algorithm*
Each test comprises 100 simulated games played. The following tests have been performed:

*4.1.1 Population vs. Generations*
Population 20 x Generation 10 vs. Population 10 x Generation 40. (Full combination is used; a total of 4000 fitness evaluations (FE) are performed for each move). The result is 82% wins for Population10x40Generations.

*4.1.2 Depth 0 vs. Depth 1*
Although the number of FE is the same, for the depth 0 the mixed chromosomes has 2 genes, for the depth 1 this number is 4. The result is 54% of wins for depth 1 (as expected).

*4.1.3 With vs. without crossover*
- The presence of crossover is fundamental for success. The result is 0 vs. 100%, favourable for with crossover.

4.1.4 Level 20 vs. Level 40 Uniform Crossover
The level is the percentage of exchanged bits for uniform crossover. The result is 57% wins for smaller percentage of exchanged bits. Too much exchange has the same effect of high random mutation.

*4.1.5 With vs. without mutation*
The presence of mutation is also fundamental. The result is 100% wins for mutation.

*4.1.6 Level 2 vs. Level 4 mutations*
The test was done between 0.2 and 0.4% bit mutation rate. The result is 38 vs. 62%, favourable top Level 4 mutation.

*4.1.7 Level 4 mutation vs. Inversion*
The inversion operator is the simple bit inversion between two random mutation points; the result is 56 vs. 44%, favourable to inversion.

*4.2 Algorithm vs. Humans*

Two 3 players groups, beginner and experienced human players (mean rating of 750) were tested. A total of 90 games are played for each group against the default CA.

*4.2.1 Default CA*
The default Ca has the following parameters:
Population = 100
Generations = 20
Crossover probability = 0.7
Uniform crossover bits % = 20
Mutation probability per bit = 0.04
Depth Level = 4

*4.2.2 Beginner*
The beginners lost all games to the default CA. A typical snapshot result is shown in figure 12.

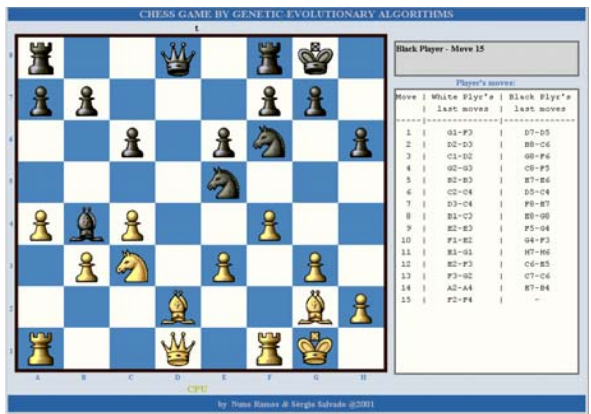
Figure 12 – Typical game between beginner and default CA.

*4.2.3 Experienced*
The result between experienced players vs. default CA is 46 vs. 54%, favourable to default CA. A typical game is shown in figure 13. It can be noted long diagonal chains of Pawns (situation of sequences of protection). The Bishops usually occupies empty diagonals, a situation that increases its influence. Castles are always performed since it is a highly rewarded move. The position of a Bishop at G2 is a very common situation, because it puts strong pressure to opponent positions

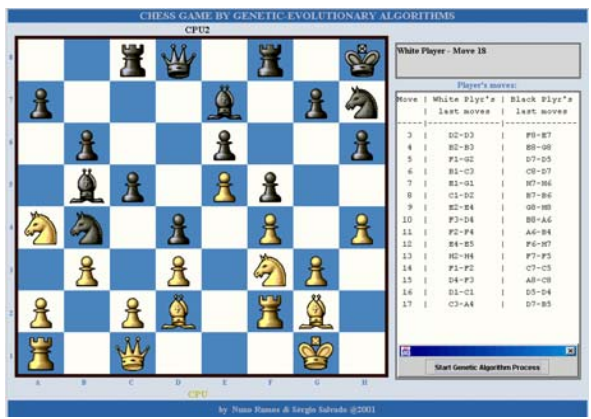
Figure 13 – Typical game between Experienced Player and default CA.

## 5 Conclusions and Discussions

A Co-evolutionary based chess player is implemented and the performance of the default CA player (that depends on the depth level) is comparable to an experienced human player.

Since finals situations are well known, they could be incorporated in order to reduce the search space. Although the scoring system used seems to work well, it has room for further improvements.

The performance of the CA player worsens in the more advanced stages of the game when the search space is much larger than in the beginning. A dynamic population and generation schedule could improve further the performance.

Currently the fitness function of the mixed chromosomes is the sum of all moves; a possibly better approach could be the fitness due to the last move in the chromosome. The final move at the specified depth is the one that matters not the intermediate moves. The danger of this strategy is the assumption that the opponent will always play the response moves coded by the simulated opponent best chromosome that is not always true.

A metalevel EA could be used to learn the weights and scorings to be used during the games and can be adapted to the opponent plat styles.

Adaptation to the international computer chess rules and platforms is under way in order to have a more precise and quantitative characterization of the CA chess Player.

## 6 References:

[1] Feng-hsiung Hsu, "Behind Deep-Blue: Building the Computer that Defeated he World Chess Champion. Princeton University Press, 2002
[2] Tang A, Moura A, Rosa AC. (1999) *"Using Genetic Algorithms in the Game Five-in-Line (Go-moku"). 2nd Int Symposium on Artificial Intelligence - Adaptive Systems - ISAS' 99, La Habana - pp:167-173.*
[3] Chellapilla K., Fogel DB (2000) *"Anaconda Defeats Hoyle 6-0 : A Case Study Competing an Evolved Checkers Program Against Commercially Available Software"*, Proc. of the Congress on Evolutionary Computing 2000, July, Vol.2, pp. 857-863.
[4] Grahan Kendall, Glenn Whitwell. (2001) *"An Evolutionary Approach for the tuning of a Chess Evaluation Function using Population Dynamics". Proc of the 2001 IEEE Congress on Evolutionary Computatutio, Seoul Korea, May, 2001, pp. 995-1002.*
[5] Pollack JB, Blair AD, Land M (1996) *"Coevolution of a Backgammon Player". Proc. of the Fifth Artificial Life Conference, May, 1996. MIT Press.*
[6] Seo YG, Cho SB, Yao X. *"Exploiting Coalition in Co-evolutionay Learning Proc. of the Congress*

*on Evolutionary Computing 2000, July, Vol.2, pp. 1268-1275.*

[7] Adelson-Velskiy, G.M., Arlazarov, V.L. and Donskoy, M.V. (1988). *Algorithms for Games*. Springer-Verlag, New York, NY. ISBN 3-540-96629-3.

[8] Ramos N, Salvado S, (2001) "*Jogos de Xadrez por Algoritmos Genético-Evolutivos*", Graduation Thesis Report, DEEC-IST-UTL 2001, http://www.laseeb.org/ChessGA.